%% file: main.tex
\documentclass{article} %
\usepackage[final]{acl}

\usepackage{times}

\input{math_commands.tex}

\input{packages_and_commands.tex}

\usepackage{graphicx}
\usepackage{hyperref}
\usepackage{url}
\usepackage{todonotes}
\usepackage{subcaption}
\usepackage{wrapfig}
\usepackage{inconsolata}
\usepackage{tabularx}
\usepackage{adjustbox}
\usepackage{listings}
\usepackage{tcolorbox}

\title{\qualeval{}: Qualitative Evaluation for Model Improvement}

\author{
Vishvak Murahari $^{\star1}$ \quad
Ameet Deshpande$^{\star1,3}$ \quad
Peter Clark$^{3}$ \quad
Tanmay Rajpurohit$^{4}$ \AND
Ashish Sabharwal$^{3}$ \quad
Karthik Narasimhan$^{1,2}$ \quad
Ashwin Kalyan$^{3}$ \AND
\normalfont{$^1$ Princeton University} \quad  \normalfont{$^2$ Princeton Language Initiative} \quad
\normalfont{$^3$ Allen Institute for AI} \quad 
\normalfont{$^4$ Independent Researcher}
\AND
\normalfont{\texttt{\{murahari,asd\}@cs.princeton.edu}}
}

\definecolor{wrong}{HTML}{FF0900}
\definecolor{maybe}{HTML}{999999}
\definecolor{right}{HTML}{0E954F}
\definecolor{inputcolor}{HTML}{FF7F00}
\definecolor{examplescolor}{HTML}{B0D7FF}
\definecolor{intsructionscolor}{HTML}{71C27D}
\definecolor{domaincolor}{HTML}{E7F2F8}
\definecolor{subtaskcolor}{HTML}{FFA384}
\begin{document}

\maketitle

\def\thefootnote{}\footnotetext{Vishvak, Ameet: Led the execution of the project and were involved in the initial ideation and vision.

Tanmay, Karthik, Ashwin: Helped with the initial ideation and vision. Provided critical guidance on concretely fleshing out the initial vision and provided valuable feedback on all aspects of the project

Peter, Ashish: Provided feedback on the structure of the initial draft}\def\thefootnote{\arabic{footnote}}

\input{sections/abstract}
\input{sections/introduction}

\input{sections/methodology.tex}

\input{sections/experimental_setup.tex}

\input{sections/results.tex}

\input{sections/analysis.tex}

\input{sections/related_works.tex}

\input{sections/conclusion.tex}

\bibliography{custom}
\newpage
\appendix
\input{sections/appendix.tex}
\end{document}

%% file: math_commands.tex
\usepackage{amsmath,amsfonts,bm,mathtools}

\def\eqref#1{equation~\ref{#1}}

\def\1{\bm{1}}

\DeclareMathAlphabet{\mathsfit}{\encodingdefault}{\sfdefault}{m}{sl}
\SetMathAlphabet{\mathsfit}{bold}{\encodingdefault}{\sfdefault}{bx}{n}

%% file: packages_and_commands.tex
\usepackage{amsmath,amsfonts}
\usepackage{booktabs,arydshln}
\usepackage{multirow}
\usepackage{ulem}
\usepackage{pifont}
\usepackage{relsize}
\usepackage{enumitem}
\usepackage{graphicx}

\newcommand{\qualevaltitle}{QualEval}
\newcommand{\qualeval}{\textsc{\qualevaltitle}}

%% file: sections/abstract.tex
\begin{abstract}
Quantitative evaluation metrics have been pivotal in gauging the advancements of AI systems like large language models (LLMs).
However, due to the intricate nature of real-world tasks, a single scalar to \textit{quantify} and \textit{compare} performance trivializes the fine-grained nuances of model behavior.
Additionally, metrics do not yield actionable diagnostics for model improvement, thus requiring extensive manual efforts of scientists, involving sifting through vast datasets and attempting hit-or-miss adjustments to training data or setups.
In this work, we address the shortcomings of quantitative metrics by proposing \qualeval{}, which uses automated \textit{qualitative} evaluation as a vehicle for model improvement.
\qualeval{} uses a powerful LLM reasoner and our novel flexible linear programming solver to generate human-readable insights that when applied, accelerate model improvement.
The insights are supported by a dashboard report with fine-grained visualizations and human-interpretable analyses.
We corroborate the faithfulness of \qualeval{} by demonstrating that leveraging its insights, for example, improves the absolute performance of the Llama 2 model by up to 15\% points relative on a challenging dialogue task (DialogSum) when compared to baselines.
\qualeval{} successfully increases the pace and quality of model development by eliminating the need of arduous manual analysis, thus serving as a data-scientist-in-a-box.

\end{abstract}

%% file: sections/introduction.tex
\section{Introduction}
\label{sec:introduction}

\input{figures-latex/teaser.tex}

The recent success of large language models (LLMs) while can be attributed to data and compute scaling, has also been the result of evaluation metrics that allow benchmarking and comparison of models.
This surge in the development of LLMs and associated tasks has reignited the need for innovative evaluation methods, aiming to provide more effective guidance throughout the model development process~\cite{tornede2023automl,paranjape2022agro}.
Traditional scalar quantitative metrics like perplexity, BLEU, and ROUGE play an important role in objectively measuring improvements in model performance.
However, these scalar metrics cannot capture the nuances of model behavior and therefore are unable to provide model developers actionable directions and diagnostics for model improvement ~\cite{DBLP:conf/emnlp/NovikovaDCR17,liu2016not}.
In practice, this deficiency necessitates model developers to collaborate with an army of data scientists and engineers to iterate over a diverse array of models and tasks, especially in rapidly evolving real-world settings.

In this work, we use ``quality over quantity'' as a guiding principle to propose our model and task agnostic method \qualeval{}, that uses qualitative evaluation to address the issues with quantitative metrics.
Given a model that is being developed for a task, \qualeval{} serves as an automated data scientist by analyzing the dataset and the model's predictions to generate actionable directions that improve the model, supported by a comprehensive dashboard containing fine-grained analysis of the model's behavior (Figure~\ref{fig:teaser}).
The directions identified by \qualeval{} to improve the model significantly expedite the model development lifecycle.
Rather than rejecting the use of metrics, \qualeval{} uses them as just one of the parts of a more holistic and useful evaluation.

\qualeval{}'s algorithm for facilitating model improvement can be broken down into three steps (Figure~\ref{fig:methodology}):
(1) \textit{Attribute discovery}: Automatic discovery of domains and sub-tasks in the dataset, to help identify issues at a fine-grained level.
(2) \textit{Attribute assignment}: Utilize a novel flexible linear programming solver to assign attributes to instances and analyze the performance on different attributes to create a human-readable dashboard.
(3) \textit{Insight generation}: Parse the generated dashboard to provide natural language insights that improve the model.
\qualeval{}'s end-to-end pipeline is completely automated and requires no human intervention.

We demonstrate~\qualeval{}'s potency on a wide range of tasks including code generation, dialogue summarization, and multiple-choice question answering.
We harness these insights provided by \qualeval{} to \textit{precisely} and significantly improve the performance of the open-source Llama 2 model on a dialog summarization task.
In a demonstration of efficacy, \qualeval{}'s insights allow a model practitioner to make changes to the fine-tuning procedure by augmenting with the \textit{right} instances, thus leading to an overall ROUGE-L score improvement of 15\%.
\qualeval{}'s insights also allow targeted improvement in sub-domains of critical importance, with absolute improvements of up to $5$ points.

Our contributions are as follows:
(1) We propose the first qualitative evaluation framework for LLMs.
(2) We introduce a novel and faithful flexible linear programming-based algorithm to automatically and accurately assign attributes to input instances, which are consequently used to generate faithful and actionable insights.
(3) We demonstrate that the generated insights can be effectively leveraged for model improvement, leading to accelerated model development.

%% file: figures-latex/teaser.tex
\begin{figure*}[t!]
    \centering
    \includegraphics[width=\textwidth]{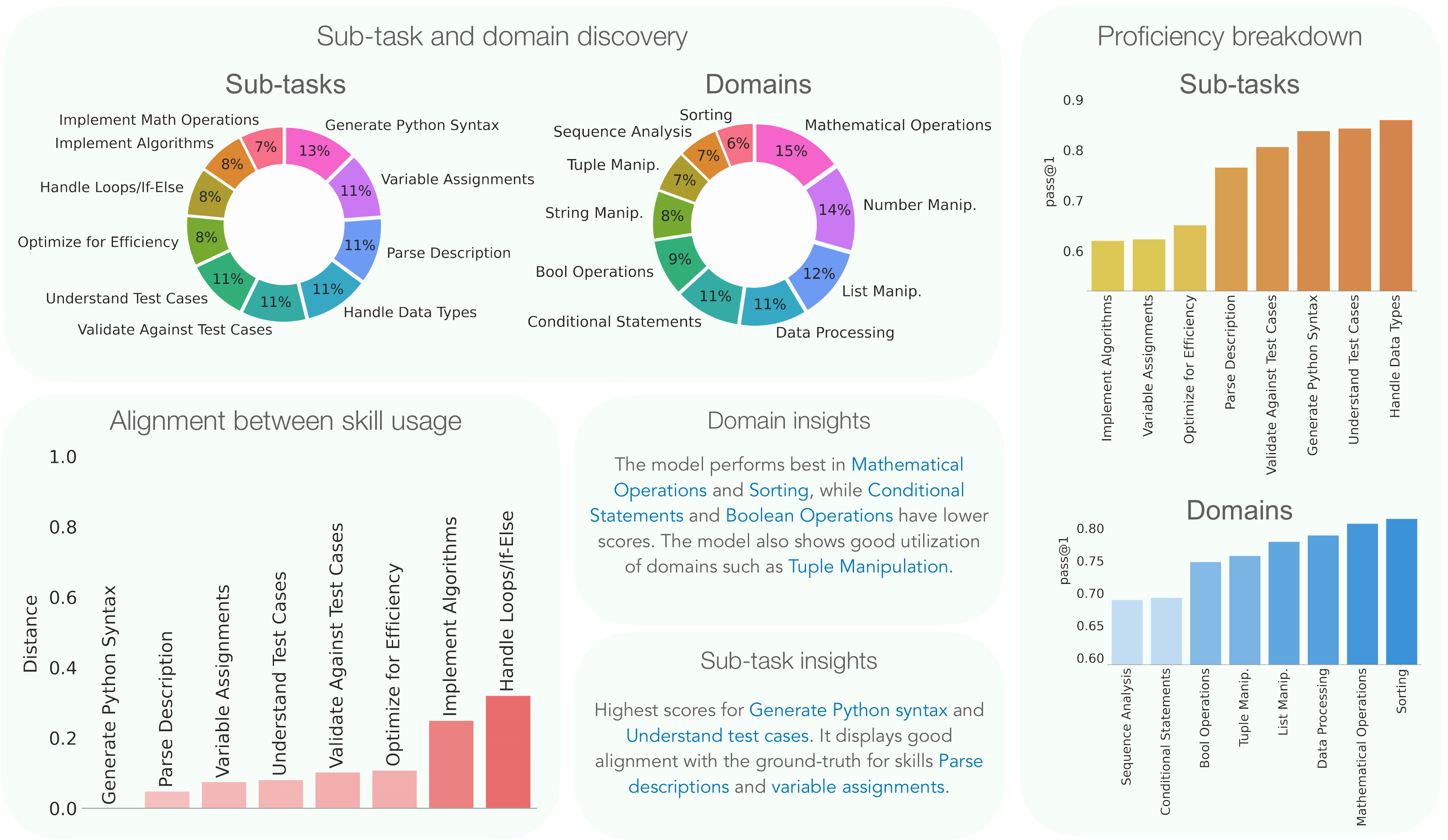}
    \caption{\qualeval{} goes beyond a single scalar metric and provides a dashboard that helps understand the model's performance in a fine-grained manner.
    \qualeval{}'s insights are faithful and lead to accelerated performance improvement when applied to the model.
    The dashboard visualizes the performance of the \textsc{davinci-3} model on MBPP.}
    \label{fig:teaser}
\end{figure*}

%% file: sections/methodology.tex
\section{Methodology}
\label{sec:methodology}

\input{figures-latex/dashboard_main.tex}

\subsection{Formulation}
\label{sec:methodology:formulation}

\paragraph{Quantitative evaluation}

\underline{Quantitative} evaluation is the standard approach to evaluating models based on averaging the value obtained by using a metric to evaluate instances of the dataset independently.
Formally, given a dataset $\mathcal{D}$ comprising of instances containing inputs ($x_i$) and ground truth outputs ($y_i$), a performance (proficiency) metric $\mathcal{M}$, and a model $f$, then:
\begin{align*}
    \mathcal{D} &\coloneqq \{(x_i, y_i)\}_{i=1}^N \\
    \hat{y}_i &\coloneqq f(x_i) \\
    \mathcal{M} &\colon (x_i, y_i, \hat{y_i}) \to \mathbb{R}\\
    \textrm{\underline{Quantitative} evaluation} &= \frac{1}{|\mathcal{D}|} \sum_{i=1}^{|\mathcal{D}|} \mathcal{M}(x_i, y_i, \hat{y_i}) \\    
\end{align*}
\vspace{-4em}
\paragraph{Qualitative evaluation}
Qualitative evaluation (\qualeval{}) is based on holistically evaluating the model's performance in a fine-grained manner rather than relying on a single scalar value.
\qualeval{} outputs a detailed dashboard that describes the intricate nature of the model's performance with the direct goal of improving it by providing actionable insights.
\qualeval{} backs up the insights with relevant evidence like visualizations and human-readable reasoning.
Formally, let $\mathcal{V}$ be the vocabulary of the language of the dashboard (here, English) and $\mathcal{I}$ be the set of all possible visualizations.
Let $\mathcal{E}$ be an evaluator system which generates the dashboard, where the system includes LLMs to reason and provide insights and image generation models to generate plots for example.
Then, the output of \qualeval{} is given by:
\begin{align*}
    \mathcal{E} &\colon \left \{ x_i, y_i, \hat{y_i} \right \}_{i=1}^N \to \left (\mathcal{V} \cup \mathcal{I} \right )^\star\\
    \textrm{\underline{Qualitative} evaluation} &= \mathcal{E}\left(\left \{ x_i, y_i, \hat{y_i} \right \}_{i=1}^N \right )\\
\end{align*}
Qualitative evaluation does not reject the use of metrics but uses them as one of the parts of a more actionable evaluation. In essence, quantitative evaluation is just a small subset of \qualeval{}.

\subsection{\qualeval{}: Qualitative evaluation}
\label{sec:methodology:qualeval}

\qualeval{} consists of multiple steps that help provide interpretable and actionable insights and we break them down below.
\paragraph{Attribute discovery}
Given the dataset $\mathcal{D}$, \qualeval{} uses an evaluator LLM ($\mathcal{E}$) to automatically discover relevant domains and sub-tasks, $d_1 \cdots d_N$ and $t_1 \cdots t_N$ in the dataset.
We refer to these domains and sub-tasks as attributes.
Specifically, we prompt $\mathcal{E}$ with the dataset and a task instruction signifying how to solve the dataset ($Instr_D$) to generate the attributes (see~\ref{appendix:prompts} for the exact prompt).
Given that datasets can have a large number of instances and LLMs have context length limits, we iteratively sample $k$ instances from the dataset and repeat the prompting process $\frac{|\mathcal{D}|}{k}$ times to generate a large list of attributes ($d_1 \cdots d_M$, $t_1 \cdots t_L$).
To ensure that we choose high-quality attributes, we prune the list of candidates in an iterative process by reducing the size by a factor of $p>1$ in each turn and repeating the process until we have $N$ attributes.
In each step, we prompt $\mathcal{E}$ to shrink the list by choosing the best attributes from the previous list of candidates.
Therefore, this iterative scalable procedure allows \qualeval{} to discover attributes in arbitrarily large data across a wide range of tasks, notwithstanding the context window limitations of $\mathcal{E}$.

\paragraph{Attribute assignment}
\qualeval{} performs attribute assignment ($d_1 \cdots d_N$ and $t_1 \cdots t_N$) by scoring the ``affinity'' or relevance of each instance with different attributes.
Let $s_{i,j}^{domain}$ and $s_{i,j}^{task}$ denote the domain and sub-task affinity scores, where $i \in \{1 \cdots |\mathcal{D}|\}$ and $j$ denotes the number of attributes ($\{1 \cdots N\}$).

We use a novel flexible linear programming solver to perform the attribute assignment by ensuring the following properties: (1) An instance is assigned $2$ domains and sub-tasks each so that we can give concrete insights.
(2) The number of assignments to an attribute is proportional to the prior probability of the attribute. This ensures that rare attributes are not ignored.
(3) Choose the assignments with maximum affinity for each instance.
We achieve the above wish-list by formulating the attribute assignment as a linear programming (LP) problem.

Given the affinity scores and the prior probabilities, $p_i$, we assign every sample to 2 domains and 2 sub-tasks.
However, we want the assignments to respect the prior probabilities i.e. ratio of the number of assignments to all the attributes should be equal to the ratio between the prior probabilities. 
We enforce this by constraining the number of assignments to an attribute to be $p_i \times |\mathcal{D}| \times 2$. 

Let $\mathbf{l}$ be the assignment matrix, where $l_{i,j} = 1$ indicates that the $i^{th}$ sample is assigned to the $j^{th}$ attribute and $l_{i,j} = 0$ indicates otherwise.
Let $p_j$ be the prior probability of the $j^{th}$ attribute.
To accommodate for the noisiness in an automated method, we make the prior probability constraint flexible by adding some slack, $\epsilon \times p_j \times |\mathcal{D}| \times 2$ ($\epsilon =0.1$) so that \qualeval{} has some wriggle room to change the attribute probability distribution in favor of better assignments.
Therefore, to enforce the prior probability constraint, we sum across the columns of $\mathbf{l}$ and constrain the sum to be between $2 \times |\mathcal{D}| \times p_j \times (1 - \epsilon)$ and $2 \times |\mathcal{D}| \times p_j \times (1 + \epsilon)$.
To ensure we assign each sample to 2 attributes, we sum across the rows of $\mathbf{l}$ and constrain the sum to be $2$.
We formalize the LP as:
\begin{small}
\begin{align*}
    \max_{\mathbf{l}} \sum_{i=1}^N \sum_{j=1}^N\: &l_{i,j} s_{i,j}^{domain/task}\\
    \sum_{j=1}^N\: &l_{i,j} = 2 \quad \forall i \in \{1 \cdots |\mathcal{D}|\}\\
    \sum_{i=1}^N\: &l_{i,j} \leq 2 * |\mathcal{D}| * p_j * (1 +  \epsilon) \quad \forall j \in \{1 \cdots N\}\\
    \sum_{i=1}^N\: &l_{i,j} \geq 2 * |\mathcal{D}| * p_j * (1 -  \epsilon) \quad \forall j \in \{1 \cdots N\}\\
    &l_{i,j} \in \{0, 1\} \quad \epsilon < 1 \quad \forall i,j \in \{1 \cdots N\}\\
\end{align*}
\end{small}
We perform an expert verification of the attribute assignments by sampling 100 samples from the dataset and asking three machine learning practitioners if both the domain and sub-task assignments are correct and find that they are indeed correct on average 84\% and 90\% of the time.

Once we have the assignments, we evaluate each instance using the proficiency metric $\mathcal{M}$ for each domain and sub-task to get $\mathcal{M}(x_i, y_i, \hat{y_i})$.
We use the assignments to breakdown the proficiency metric by domains and sub-tasks and automatically generate visualizations that help understand the model's fine-grained performance.

\paragraph{Measuring sub-task skill alignment}
For several datasets, predicting the right answer is not good enough, and producing an answer that uses the same sub-tasks as the ground truth is important.
We call this skill alignment and compute it by measuring the correlation between the sub-task affinity scores of the ground truth and the model prediction (higher values implying higher skill alignment).

\paragraph{Insight generation}
\qualeval{} then leverages the visualizations from previous stages to generate useful and actionable insights as a natural language output.
We prompt $\mathcal{E}$ with the data from the prior probability, proficiency breakdown, and skill alignment visualizations to generate useful insights (See~\ref{appendix:prompts} for exact prompt).
We integrate all the visualizations and insights into a human-readable dashboard depicted in Figure~\ref{fig:teaser}.

%% file: figures-latex/dashboard_main.tex
\begin{figure*}[t!]
    \centering
    \includegraphics[width=\textwidth]{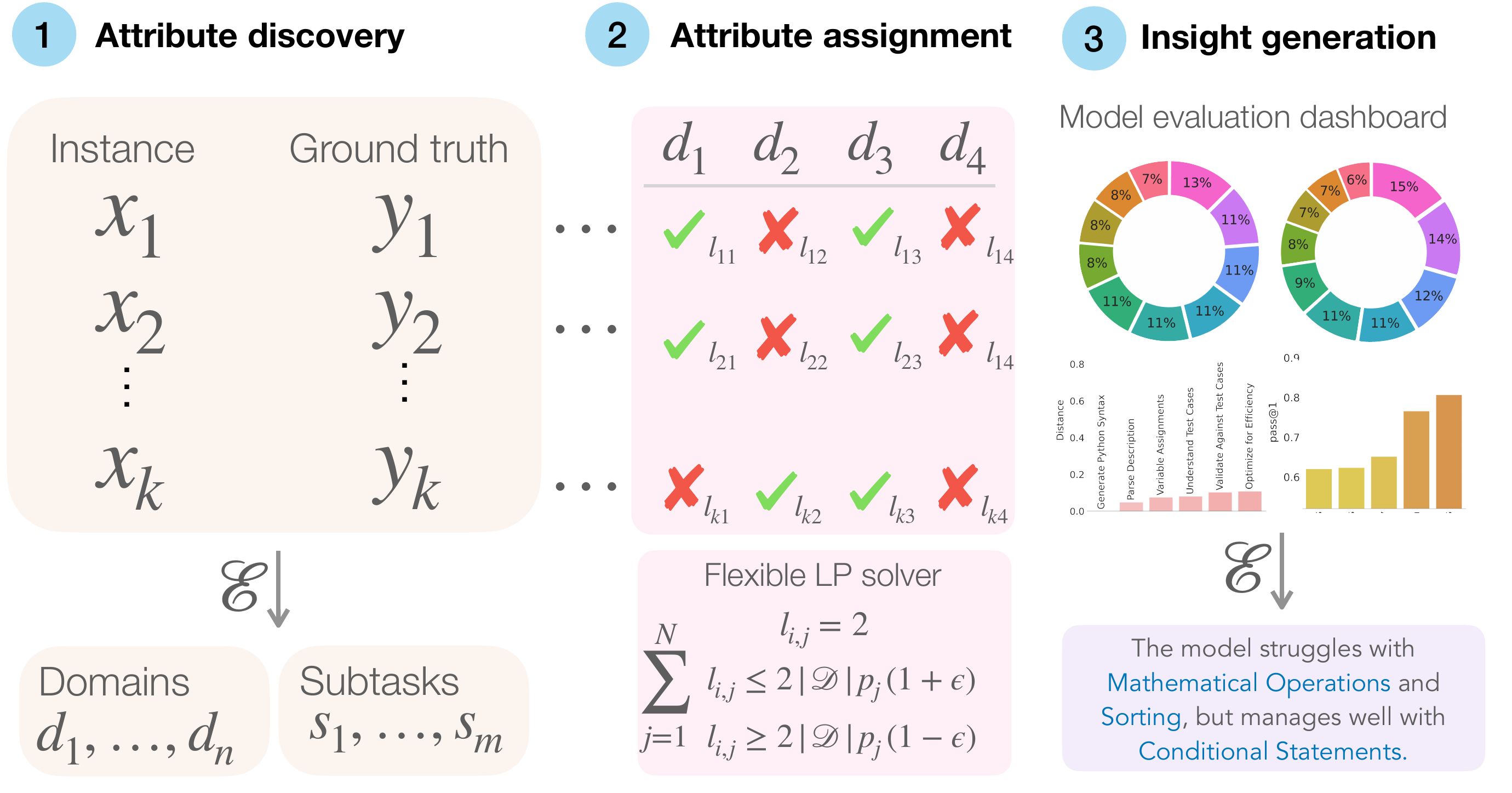}
    \caption{\qualeval{} automatically discovers domains and sub-tasks from input data through an evaluator LLM, $\mathcal{E}$. \qualeval{} then automatically assigns 2 domains and 2 sub-tasks to every sample in the dataset by solving a flexible linear program. Finally, \qualeval{} generates a comprehensive dashboard and presents interpretable and actionable insights for practitioners.
    } 
    \label{fig:methodology}
\end{figure*}

%% file: sections/experimental_setup.tex
\section{Experimental Setup}
\paragraph{Datasets}
We evalaute \qualeval{} on three datasets: MBPP~\citep{mbpp} (sanitized), DialogSum~\citep{dialogsum}, and MMLU~\citep{mmlu} (clinical knowledge split). MBPP and DialogSum involve generative tasks and involve generating a Python program from a prompt and summarizing a conversation respectively. MMLU contains a wide range of multiple-choice questions from different domains and we pick the clinical knowledge split to evaluate our model on knowledge-intensive tasks.
We use the same evaluation splits as the original papers and use the test splits for MBPP and MMLU and the validation split for DialogSum. We use the pass@1, ROUGE-L, and accuracy as proficiency metrics for MBPP, DialogSum, and MMLU respectively.
\paragraph{Models}
We use both closed and open-sourced models: \textsc{curie}, \textsc{davinci-2}, and \textsc{davinci-3} models from OpenAI and Llama 2 (7 billion chat models~\citep{llama2}).
We use a temperature of $0.9$ for all models and use two randomly sampled in-context samples for prompting models unless mentioned otherwise.
We instantiate $\mathcal{E}$ with the \textsc{gpt-3.5-turbo} model~\citep{chatgpt}.
\paragraph{Llama fine-tuning} 
We use LoRA~\cite{lora} to efficiently fine-tune the Llama 2 7 billion parameter model and train with $8$ bit precision.
We sweep over five learning rates ($2e-5$, $5e-5$, $1e-4$, $2e-4$, $1e-3$) and pick the checkpoint with the best validation performance. 
We train for up to $400$ steps and we use a constant learning rate schedule.
\paragraph{Attribute generation}
We set N (the initial number of generated categories) to 15, p (the pruning factor) to 4, and k (the number of few-shot examples during category generation) to 5 in our experiments.

%% file: sections/results.tex
\section{Results}

We systematically present different aspects of our dashboard.
Firstly, we show that attribute discovery (domains and sub-tasks) of \qualeval{} is well-grounded and faithful to the dataset.
Secondly, we show that \qualeval{}'s flexible LP solver correctly assigns attributes to instances of the dataset, allowing it to perform meta-reasoning over different domains and sub-tasks.
Finally, we validate that the concise natural language insight generated leads to improvement in the model's performance.

\subsection{Discovering Domains and Sub-tasks}
\input{figures-latex/prior_dashboard_multiple_tasks.tex}

Discovering the latent domains and sub-tasks in a dataset and understanding their prominence through the prior probability of their occurrence is a critical step for \qualeval{}.
\qualeval{} performs both the discovery and prior-probability computation \textit{automatically} and \textit{faithfully}.

As an example, Figure~\ref{fig:prior_viz} presents the prior probabilities of the domains and sub-tasks in the MBPP and DialogSum datasets.
We find that the MBPP dataset comprises a large set of samples that involve domains like mathematical/numerical operations ($29\%$) and list manipulation ($12\%$) while domains like sorting ($6\%$) and tuple manipulation ($7\%$) are less prevalent.
Interestingly, \qualeval{} captures fine-grained nuances by including closely related yet different sub-tasks like ``Implement mathematical operations'' and ``Implement algorithmic operations'', giving practitioners a nuanced understanding of their evaluation data.

As another illustration (Figure~\ref{fig:prior_viz} bottom), the DialogSum dataset is dominated by samples involving domains like employment and professional skills ($15\%$) and career and job interviews ($14\%$), while domains like education and outdoor activities are less prevalent ($8\%$ and $8\%$ respectively).
Though the overall food domain is also frequent, it is listed under two fine-grained domains, ``Food and restaurant ordering'' ($7\%$) and ``Food and hospitality'' ($8\%$), which further highlights \qualeval{}'s ability to capture fine-grained nuances.
The evaluation also suggests the dominance of sub-tasks that involve identifying the participants ($12\%$), understanding and recognizing the main topic ($22\%$ ), and recognizing the roles in the conversation ($11\%$), which are conceptually important sub-tasks for summarizing a conversation between two people.

\paragraph{Faithfulness of priors}
\input{figures-latex/medmcqa_prior.tex}
We show that the attributes discovered and prior probabilities assigned are faithful to the dataset.
While most datasets do not have ground truth annotations for the domains and sub-tasks,~\cite{medmcqa} introduces a multiple-choice question answering dataset, MedMCQA, collected from real-world medical exam questions, and includes domain annotations.
We randomly sample $250$ questions from the MedMCQA dataset and leverage \qualeval{} to discover domains and find the prior probabilities.
We compare the prior probabilities from \qualeval{} with the ground truth domain annotations from MedMCQA in Figure~\ref{fig:medmcqa_prior}.
We find that the domain priors from \qualeval{} are highly aligned with the ground truth annotations (``Pediatrics'' ($9\%$ vs $9\%$), ``Obstetrics and Gynecology''($6\%$ vs $7\%$), and ``Pharmacology'' ($6\%$ vs $6\%$) and ``Microbiology'' ($4\%$ vs $6\%$)).
Interestingly, \qualeval{} splits the ``Dental'' domain into more precise domains such ``Dentistry'', ``Dental Hygiene'', ``Dental procedures'', and ``Dental anatomy'', further highlighting \qualeval{}'s ability to capture hierarchies and nuances in the data.

\subsection{Proficiency categorized by Domains and sub-tasks}
\input{figures-latex/proficiency_dashboard_multiple_tasks.tex}

To generate useful insights, one needs a clear understanding of the model's proficiency in the various domains and sub-tasks, and we demonstrate that \qualeval{} provides exactly this. \qualeval{} leverages the domain and sub-tasks assignments generated from our flexible LP solver to get a precise breakdown of the proficiency of a model for different domains and sub-tasks.

Figure~\ref{fig:proficiency_viz} highlights the proficiency of the \textsc{davinci-3} model on domains like sorting, mathematical operations, and data processing and on sub-tasks like handling data types, understanding test cases, and generating Python syntax.
We find that \qualeval{}'s categorization and proficiency judgement is faithful and aligned, as corroborated by analysis from~\citet{mbpp} that also suggests that models on MBPP perform well on ``coding interview'' type questions which generally involve data structures, sorting, list manipulation, and data processing.

~\citet{mbpp} also suggests that models struggle with samples related to advanced math problems and samples with multiple sub-problems.
This conforms with \qualeval{}'s proficiency breakdown which reveals that the model struggles with samples involving the ``Implement algorithms'' and ``Variable assignments'' sub-tasks and the ``Conditional statements'' and ``Sequence Analysis'' domains, which are often leveraged to solve math problems and samples with multiple sub-problems. These findings serve to reinforce the distinctive capability of \qualeval{} in offering a precise and nuanced comprehension of model proficiency, made possible by our flexible LP solver.

\qualeval{} is task-agnostic, with our flexible LP solver making it potent even in niche domains such as clinical data.
Figure~\ref{fig:proficiency_viz} demonstrates high proficiency of the \textsc{davinci-3} model on the cell biology and medical procedures domains and sub-tasks related to analyzing and processing medical test data and recognizing key terms/concepts in clinical biology.
However, the model struggles with sub-tasks related to providing accurate information and analyzing the correct answer choice.

\subsection{Interpretable and Actionable natural language insights}
To aid model developers in understanding the dense fine-grained analysis in the prior sections, we present interpretable and actionable natural language insights grounded in the prior analysis.
To generate these insights, we convert the analysis charts depicted in the prior sections into structured text and query our evaluator LLM to highlight important and actionable trends and insights.
Figure~\ref{fig:insights} illustrates insights generated by \qualeval{} for predictions from \textsc{davinci-3} model on MBPP, which adeptly highlights model deficiencies for both domains and sub-tasks.
For instance, the insights faithfully point out that improvements can be made to the ``Tuple Manipulation'' and ``Number Manipulation'' domains as well as the ``Algorithmic operations'' and ``Handling loops and conditionals'' sub-tasks.
In the next section, we demonstrate how these insights can be leveraged towards precise and targeted model improvement, further validating the efficacy of \qualeval{}.

\input{Tables/table_model_improvement}
\subsection{Model Improvement via Qualitative Evaluation}
\label{sec:model_improvement}

We show that \qualeval{}'s actionable insights are useful by improving models on a variety of settings on the DialogSum dataset.
We leverage insights from \qualeval{} to \textit{precisely} and \textit{consistently} improve the proficiency of a 7 billion parameter Llama 2 model.

Consider a real-world scenario where certain sub-domains are more important.
For example, in a toxicity detection dataset, you would expect sub-domains relating to \textit{racial abuse} to have better accuracy than say \textit{politics}.
In such a case, a practitioner would want to identify if there are critical sub-domains where the model under-performs, and fix those issues.
We consider this scenario for the Llama model where the practitioner is allowed to augment a certain number of instances of their choice to the training set based on the insights.
This simulates the scenario where only a certain number of annotated examples can be obtained because of data paucity and cost reasons.

Assume there is a set of sub-domains that the model is underperforming in, as identified by \qualeval{}.
\qualeval{}'s flexible LP solver finds a set of unsupervised examples belonging to these domains that are then annotated and added to the training instances.
We compare with a baseline (Rand. Aug.) that randomly annotates and augments the same number of instances from the unsupervised store.
We experiment with different sets of under-performing domains in Table~\ref{tab:model_improvement} (pertaining to different rows) by fine-tuning the Llama 2 model on the two augmented dataset settings.
Additional details are presented in Appendix~\ref{app:modelimprovement}.

Across different sets of domains (rows), \qualeval{} consistently and significantly increases the proficiency of the selected domains and the overall performance (Table~\ref{tab:model_improvement}).
For instance, augmenting with the  ``Leisure'', ``Food ordering'', and ``Hospitality'' domains (last row) leads to an improvement of $5.4\%$, $4.2\%$, and $2.9\%$ on ROUGE-L on the respective domains and an overall improvement of $4.1\%$ on ROUGE-L, when compared to \textit{Rand. Aug}.
Taken together, \qualeval{} empowers practitioners to improve model proficiency with a high degree of \textit{precision} and \textit{control}.

%% file: figures-latex/prior_dashboard_multiple_tasks.tex
\begin{figure*}[t!]
    \centering
    \begin{subfigure}{\textwidth}
        \centering
        \includegraphics[width=\textwidth]{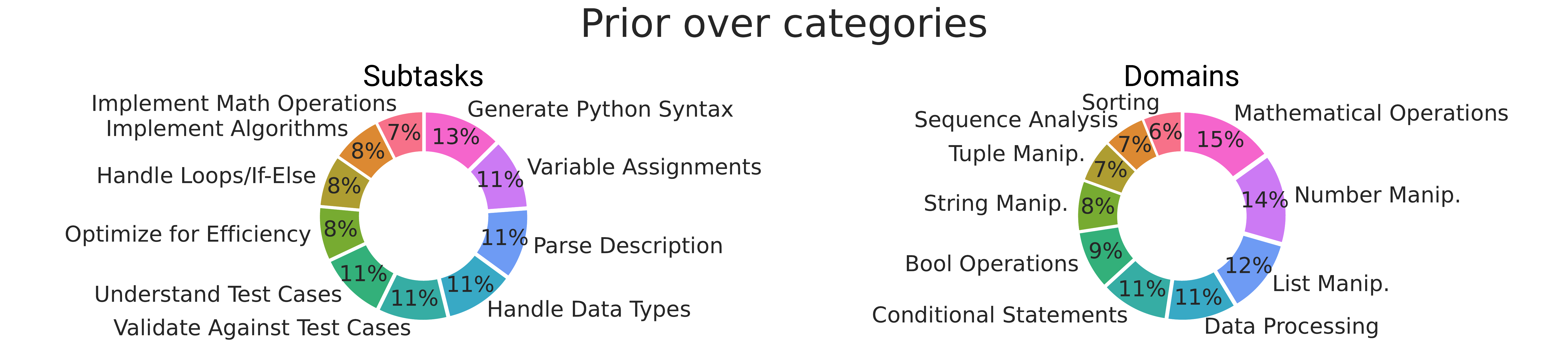}
    \end{subfigure}
    \begin{subfigure}{\textwidth}
        \centering
        \includegraphics[width=\textwidth]{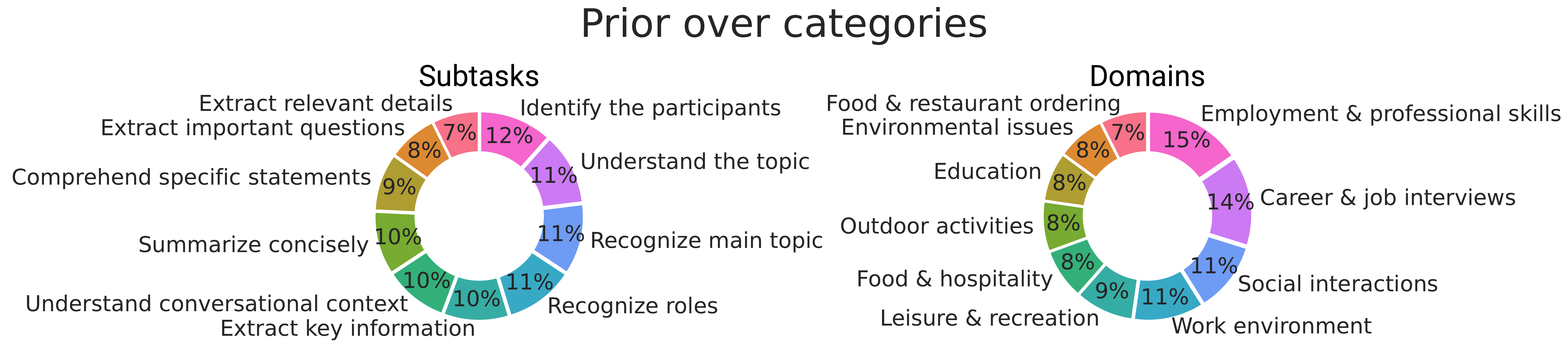}
    \end{subfigure}
    \caption{Prior probabilities of domains and sub-tasks on the MBPP (top) and DialogSum (bottom) datasets}
    \label{fig:prior_viz}
\end{figure*}

%% file: figures-latex/medmcqa_prior.tex
\begin{figure*}
    \centering
    \includegraphics[width=\textwidth]{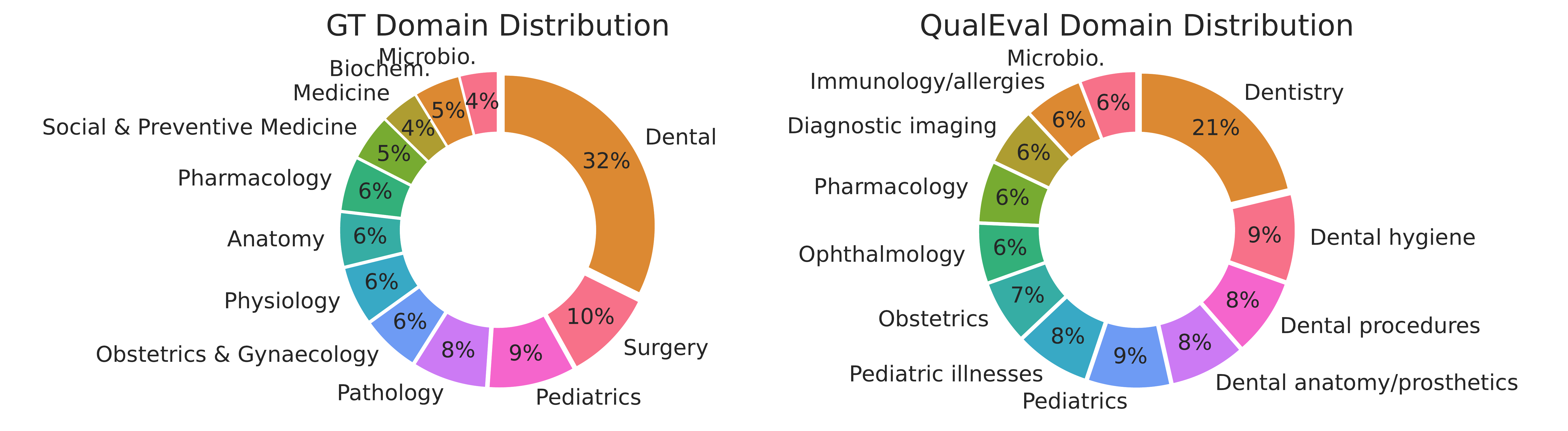}
    \caption{\qualeval{} faithfully discovers and scores attributes.
    We compare the domain priors discovered by \qualeval{}(right) with the ground truth domain annotations (left) in the MedMCQA dataset and find a high degree of alignment (e.g., ``Pediatrics'' -- $9\%$ vs $9\%$, ``Obstetrics and Gynecology'' -- $6\%$ vs $7\%$, and ``Pharmacology'' -- $6\%$ vs $6\%$).
    }
    \label{fig:medmcqa_prior}
\end{figure*}

%% file: figures-latex/proficiency_dashboard_multiple_tasks.tex
\begin{figure*}[t!]
    \centering
    \begin{subfigure}{0.48\textwidth}
        \centering
        \includegraphics[width=\textwidth]{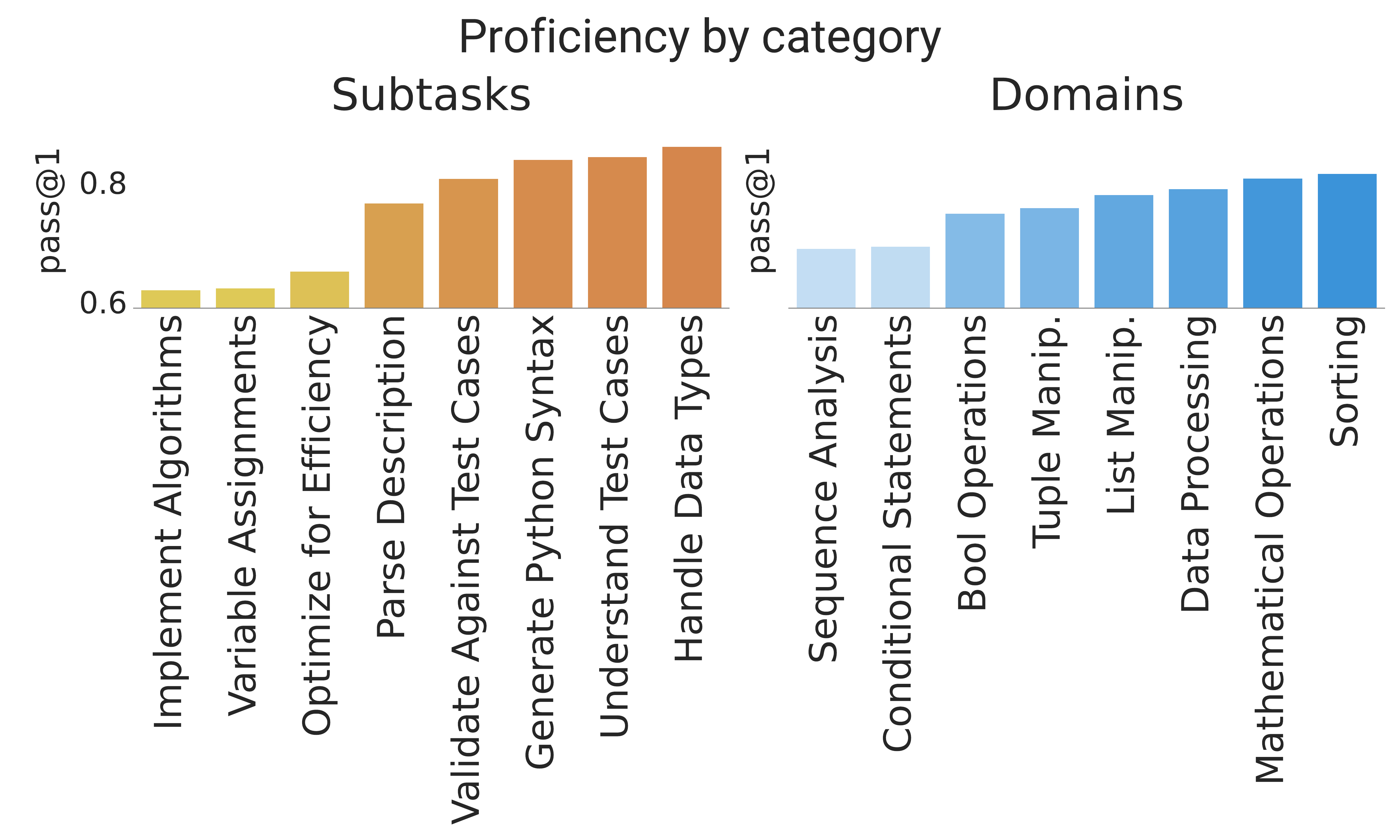}
    \end{subfigure}
    \begin{subfigure}{0.48\textwidth}
        \centering
        \includegraphics[width=\textwidth]{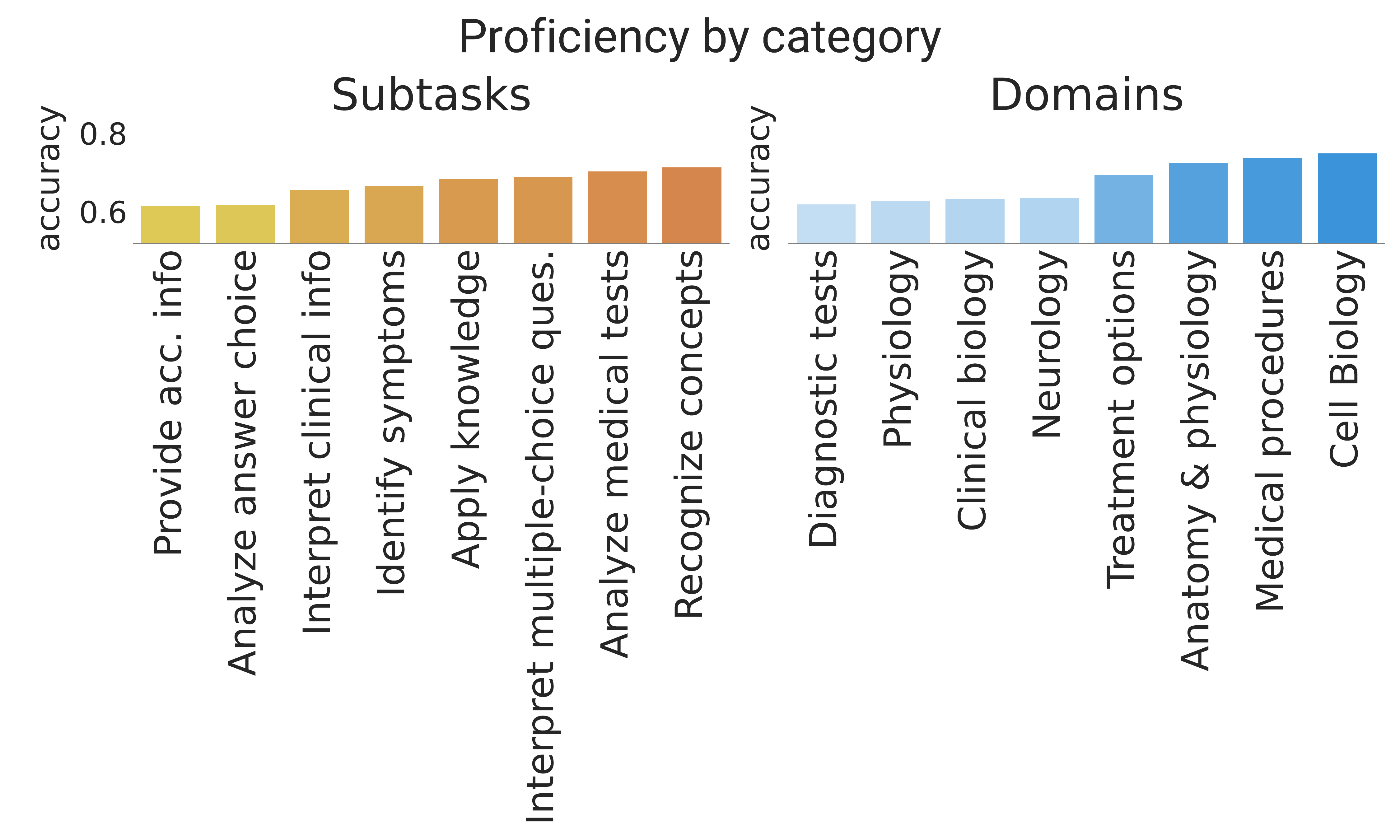}
    \end{subfigure}
    \caption{Proficiency breakdown for different sub-tasks and domains in the MBPP and MMLU (clinical knowledge) datasets for \textsc{davinci-3}.
    }
    \label{fig:proficiency_viz}
\end{figure*}

%% file: Tables/table_model_improvement.tex
\newcommand{\nonumbertable}{\color{gray}\ding{55}\color{black}}

\makeatletter
\def\adl@drawiv#1#2#3{%
        \hskip.5\tabcolsep
        \xleaders#3{#2.5\@tempdimb #1{1}#2.5\@tempdimb}%
                #2\z@ plus1fil minus1fil\relax
        \hskip.5\tabcolsep}
\newcommand{\cdashlinelr}[1]{%
  \noalign{\vskip\aboverulesep
           \global\let\@dashdrawstore\adl@draw
           \global\let\adl@draw\adl@drawiv}
  \cdashline{#1}
  \noalign{\global\let\adl@draw\@dashdrawstore
           \vskip\belowrulesep}}
\makeatother
\setlength{\dashlinedash}{0.30\dashlinedash}

\begin{table*}[t]
\centering
\resizebox{\textwidth}{!}{%
\begin{tabular}{lcccccccccccc}
\toprule
\multicolumn{3}{c}{\textbf{Domain sets}} & \multicolumn{3}{c}{\textbf{Rand. aug.}} & \multicolumn{3}{c}{\textbf{\qualeval{} aug.}} &\multicolumn{4}{c}{{\small $\mathbf{\Delta}$ = \textbf{(\qualeval{} aug. -- Rand. aug.)}$\uparrow$}} \\
\cmidrule(lr){1-3} \cmidrule(lr){4-6} \cmidrule(lr){7-9} \cmidrule(lr){10-13}
Dom 1 & Dom 2 & Dom 3 &  Dom 1 & Dom 2 & Dom 3  & Dom 1 & Dom 2 &  Dom 3 & Dom 1 & Dom 2 & Dom 3 & Overall \\ \midrule
Social & \nonumbertable & \nonumbertable & $27.6$ & \nonumbertable & \nonumbertable & $30.0$ & \nonumbertable & \nonumbertable & $\mathbf{2.4}$ & \nonumbertable & \nonumbertable & $\mathbf{2.6}$  \\ \cdashlinelr{1-13}
Leisure & Outdoor & \nonumbertable & $26.6$ & $27.1$ & \nonumbertable &  $29.0$ & $27.7$ & \nonumbertable & $\mathbf{2.4}$ & $\mathbf{0.6}$ & \nonumbertable & $\mathbf{3.1}$  \\ \cdashlinelr{1-13}
Food ordering & Hospitality & \nonumbertable &  $27.8$ & $28.3$ & \nonumbertable & $31.5$ & $31.5$ & \nonumbertable & $\mathbf{3.7}$ & $\mathbf{3.2}$ & \nonumbertable & $\mathbf{3.6}$ \\ \cdashlinelr{1-13}
Leisure & Food Ordering & Hospitality & $26.6$ & $27.8$ & $28.3$ & $32.0$ & $32.0$ & $31.2$ & $\mathbf{5.4}$ & $\mathbf{4.2}$ & $\mathbf{2.9}$ & $\mathbf{4.1}$ \\
\midrule
\end{tabular}
}
\caption{\qualeval{} consistently increases the performance (ROUGE-L) of the Llama 2 (7 billion parameter) model on DialogSum.
\qualeval{} enables practitioners to do targeted model improvement through data augmentation, while keeping the training set size constant.
We demonstrate improvements across different sets of domains (with different domains and different numbers of domains) and show consistent and significant improvements on the selected domains along with improvements in overall performance (refer to columns under ``$\Delta$'').
For instance, augmenting with the  ``Leisure'', ``Food ordering'', and ``Hospitality'' domains (last row) leads to an \textit{absolute} overall improvement of $4.1$ percentage points.
}

\label{tab:model_improvement}
\vspace{-10pt}
\end{table*}

%% file: sections/analysis.tex
\section{Analysis}
\label{sec:analysis}

\subsection{Skill usage calibration between ground truth and model answers}
\input{figures-latex/corr_dashboard_multiple_tasks.tex}
While proficiency metrics like pass@k, BLEU, and ROUGE can judge the proficiency of a model,
they do not provide insights about skill usage calibration, i.e., whether the model is leveraging the expected subtasks when generating responses.
Skill usage calibration is a unique lens to understand model performance, as practitioners can understand if the model generates answers with the expected and intended reasoning.

We quantify the calibration by first identifying the affinity of the ground truth and model-generated answers to different sub-tasks discovered.
We then measure the distance between the affinity scores.
A smaller distance implies that the model-generated answer is using sub-tasks similar to the ground truth answer, thus exhibiting high skill usage calibration.
We explain the exact distance metric used in Appendix~\ref{app:distance}.

Figure~\ref{fig:corr_viz} highlights the correlation between model generations and ground truth responses for the \textsc{davinci-3} model on the MBPP and DialogSum datasets.
A model practitioner can utilize this to understand what sub-tasks are not being used in an intended way and perform an intervention to fix them.
For example, on the MBPP dataset, the subtasks about implementing algorithms and handling loops and conditionals have low alignment.

\subsection{Qualitative Samples}
\label{analysis:qualitative}
\input{figures-latex/mbpp_qualsamples.tex}

\qualeval{} also allows model developers to extract prominent qualitative examples that can aid in the modeling lifecycle.
Given that both in an academic and industry setting, understanding representative instances of ground truth and model-generated answers is important, \qualeval{} automates that process.
It \textit{automatically} yields revealing qualitative samples by identifying samples where the affinity scores of the ground truth response and model generation are not aligned.

Figure~\ref{fig:mbpp_qualitative_samples} shows qualitative samples from the MBPP dataset generated by the \textsc{davinci-3} (left and center) and \textsc{davinci-2} (right) models.
In the first example, the ground truth program uses XOR to test for uniqueness, while the generation uses a loop to check for uniqueness.
In the second example, the ground truth program uses an in-built Python function to check equality whereas the model loops through the input to check the condition.
These examples further validate the finding in the prior section which suggests that the model is not calibrated for handling loops and conditionals.

Interestingly, the generated output in the final example is a more robust solution than the ground truth.
The ground truth solution assumes that the input is a list of booleans, while the model generation can accept any list with any data type.

%% file: figures-latex/corr_dashboard_multiple_tasks.tex
\begin{figure}[h!]
    \centering
    \begin{subfigure}{\linewidth}
        \centering
        \includegraphics[width=0.75\textwidth]{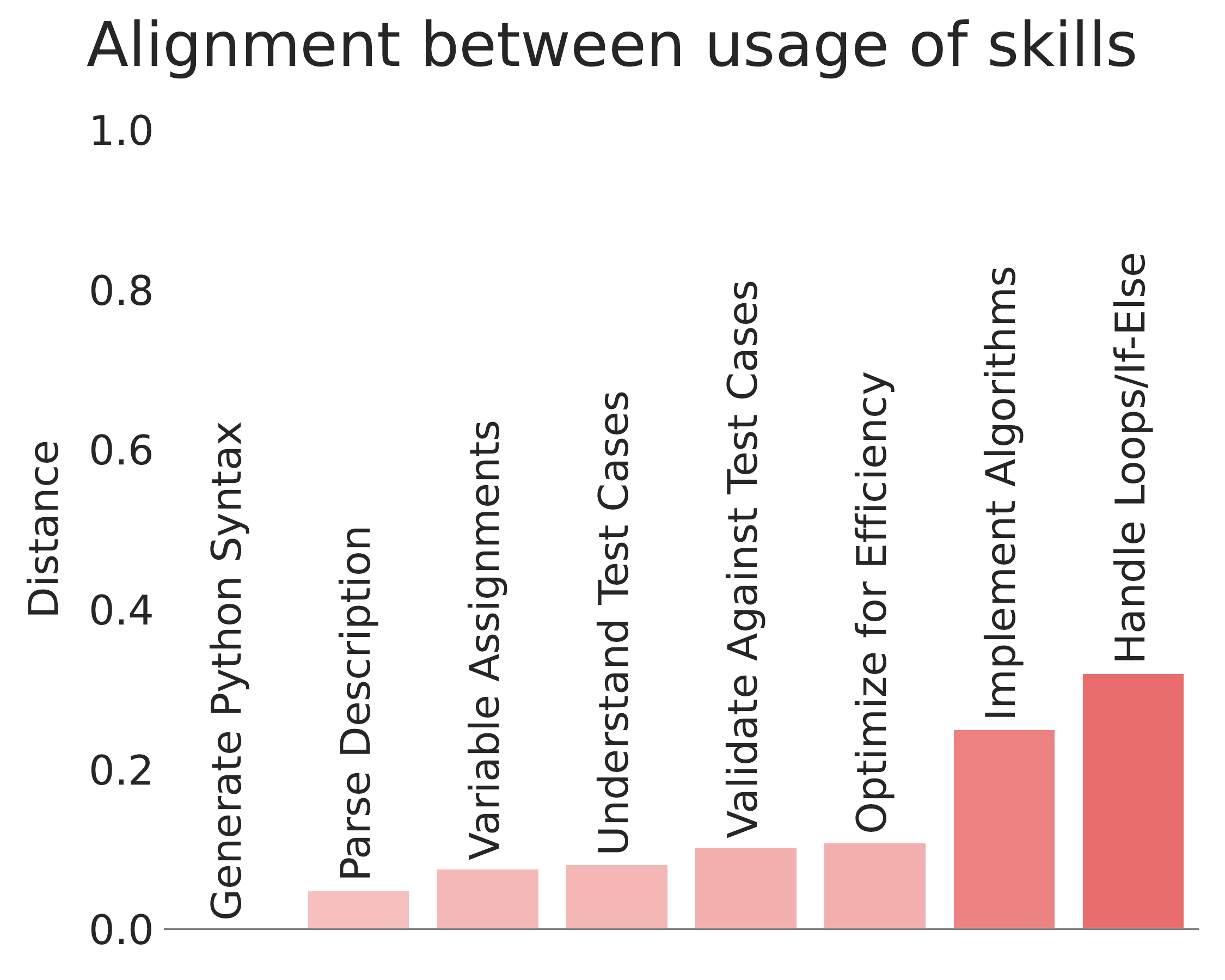}
        \caption{}
        \label{fig:mbpp_dist}
    \end{subfigure}
    \begin{subfigure}{\linewidth}
        \centering
        \includegraphics[width=0.75\textwidth]{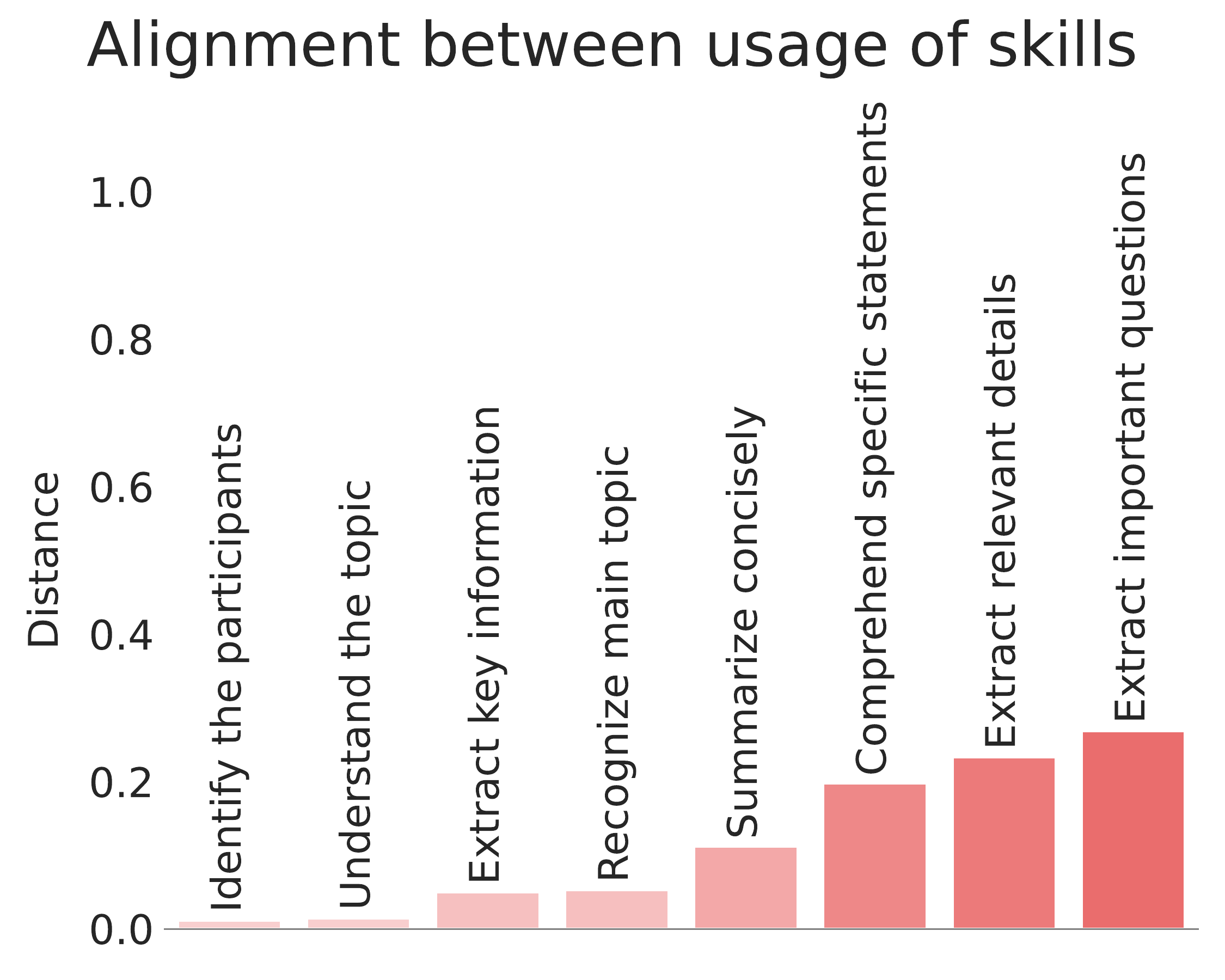}
        \caption{}
        \label{fig:dialogsum_dist}
    \end{subfigure}
    \caption{Skill usage calibration between the ground truth and the model generated answer for the \textsc{davinci-3} model on MBPP (top) and DialogSum (down). A smaller distance implies that the model is using sub-tasks as intended in the ground truth (hence it is better).}
    \label{fig:corr_viz}
\end{figure}

%% file: figures-latex/mbpp_qualsamples.tex
\begin{figure}[htp]
    \centering
    \includegraphics[trim={0 2cm 6cm 0},clip,width=\linewidth]{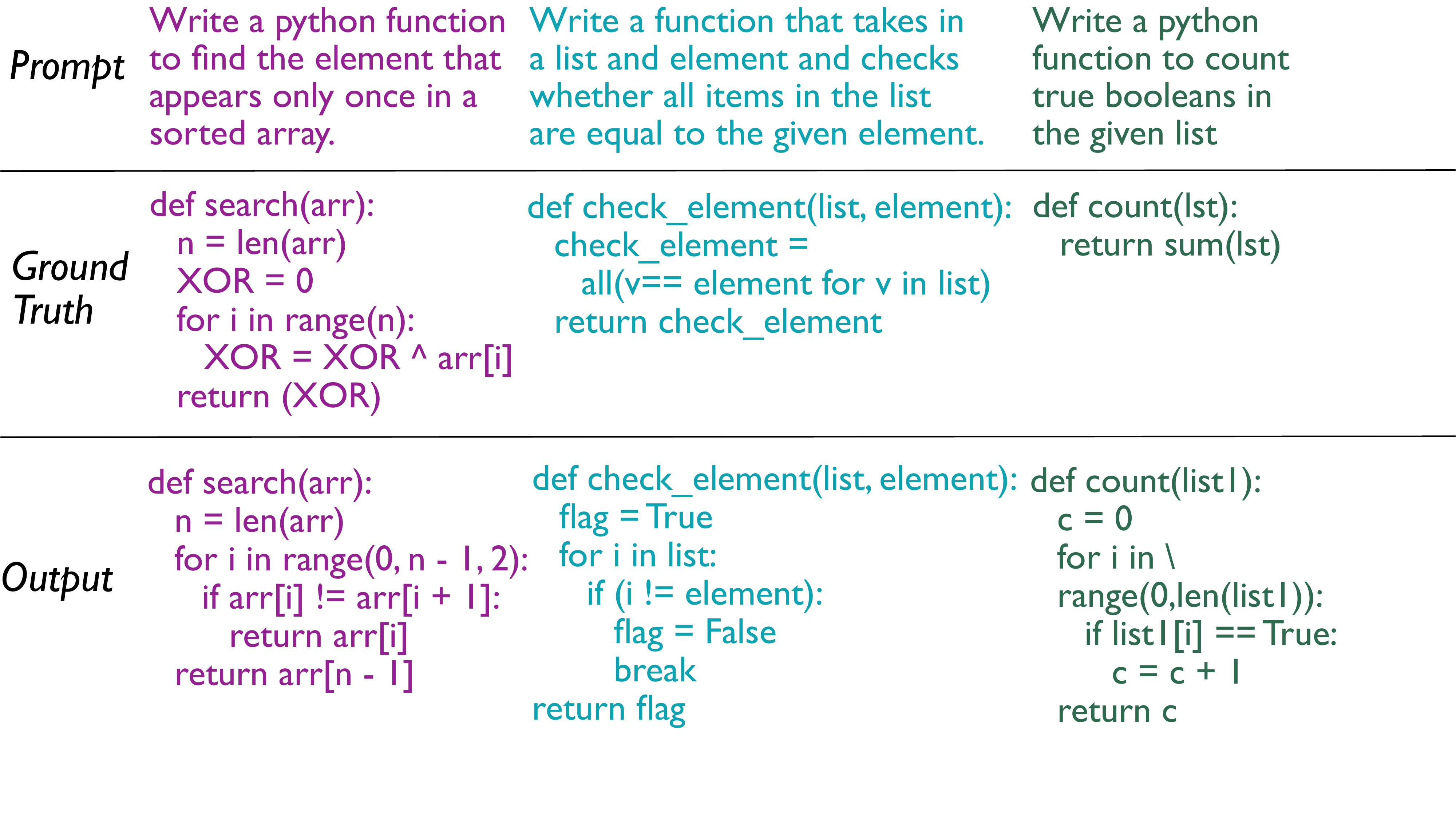}
    \caption{Qualitative samples from the MBPP dataset retrieved through \qualeval{}.
    We provide an explanation for their prominence in Section~\ref{analysis:qualitative}.
    }
    \label{fig:mbpp_qualitative_samples}
\end{figure}

%% file: sections/related_works.tex
\section{Related Work}

\paragraph{Model Debugging/Improvement}
Prior work has attempted to address the problem of model debugging and improvement.
\cite{zhang2018manifold} propose to evaluate different pairs of models on separate evaluation splits to understand model behavior.
They also generate feature-level importance scores from ``symptom'' instances provided by humans.
\cite{gralinski-etal-2019-geval} introduce a model-agnostic method to find global features that ``influence'' the model evaluation score, allowing practitioners to exclude problematic features.
\cite{lertvittayakumjorn-toni-2021-explanation} develop a framework to generate explanations for model predictions to allow humans to give feedback and debug models.
\cite{checklist} presents a framework to generate test cases at scale to evaluate model robustness, but constrains the test cases to be generated from simple templates and lexical transformations. 
\cite{pmlr-v162-abid22a} propose a framework to generate counterfactual explanations for model errors to enable a better understanding of model behavior.
\cite{selfdebug} introduce Self-Debugging, a method to enable a large language model to debug the predicted computer program through few-shot demonstrations.
Some other works attempt to find error-prone slices of the data to improve the model~\cite{he2021automl,tornede2023automl,paranjape2022agro}.
While these works provide limited insights into model behavior, they often require significant human intervention to understand model behavior and do not provide precise actionable insights for model improvement.
Finally, these works are constrained to simple classification and regression tasks or single domains like code generation and do not provide a task-agnostic, fully automated framework for model interpretation and improvement for real-world tasks.

\paragraph{Automatic Evaluation of Machine Learning Models}
Automatic evaluation metrics, based on lexical overlap, such as BLEU~\cite{bleu}, ROUGE~\cite{rouge}, METEOR~\cite{meteor} have helped researchers evaluate and compare models on a variety of language tasks.
Recent work has proposed to use machine learning models to evaluate other machine learning models. Methods like ~\cite{bertscore, gptscore, codebertscore} use pre-trained language models to evaluate the quality of generated text and therefore rely more on semantics than lexical overlap.
While these automated metrics have expedited research progress by eliminating human effort from evaluation, they have limited evaluation to a single scalar metric and therefore fail to provide a holistic and comprehensive understanding of model performance.

\paragraph{Issues with quantitative metrics}
Multiple studies have pointed out that quantitative metrics are not sufficient to understand the behavior of LLMs and that they are not a good proxy for real-world performance~\cite{liu2008correlation,DBLP:conf/emnlp/NovikovaDCR17,reiter2009investigation,liu2016not}.
While these studies advocate better quantitative metrics, our study proposes a new framework based on \textit{qualitative} evaluation.

%% file: sections/conclusion.tex
\section{Conclusion}
\label{sec:conclusion}

We propose \qualeval{}, a qualitative evaluation framework that provides a comprehensive way of evaluating models with a keen eye on model improvement.
Rather than rely on scalar quantitative metrics that ignore the nuanced behavior of the model, \qualeval{} augments quantitative metrics to test the model thoroughly and provides actionable insights through an interpretable dashboard to improve the model iteratively.
We demonstrate that these insights are faithful and lead to up to 15\% relative improvement.
Our work is the first step towards building a data-scientist in a box.

\section{Limitations}
\qualeval{} can faithfully discover relevant sub-tasks and domains and can generate interpretable and actionable dashboards from model predictions. 
However, we only demonstrate \qualeval{} on diverse language tasks like code generation, summarization, and question answering but do not demonstrate results on multi-lingual and multi-modal tasks.
Our evaluator model is a closed-source model (\textsc{gpt-3.5}), and replacing it with a performant open-source model will make our contribution more accessible to the community. We hope future work will address these challenges and extend \qualeval{} to be an even more general paradigm.

\section{Ethical Considerations}
\label{sec:ethics}

Our work provides a potent way to ensure that certain tasks performed by data scientists can be automated.
While this reduces the burden on them, it is also possible that it reduces the need to have a very large group of them on a certain project.
This might have workforce implications.
But the intention of the study is to show that with the current LLMs, we can improve evaluation by making it comprehensive.

%% file: sections/appendix.tex
\section{Appendix}

\subsection{Prompts used in \qualeval{}}
\input{figures-latex/discovery_prompt}
\label{appendix:prompts}

\input{figures-latex/scoring_prompt}
\input{figures-latex/insight_prompt}

\subsection{Model improvement}
\label{app:modelimprovement}

We first leverage \qualeval{}'s flexible LP solver to generate domain assignments for training samples.
We then choose a base set of $250$ training samples and leverage the domain assignments to augment the training set by adding $250$ additional samples from the training set from up to $3$ different domains.
Therefore we randomly sample $\frac{250}{N}$ samples from the selected domains, where $N$ is the number of selected domains ($N\leq3$).
We experiment with different sets of domains in Table~\ref{tab:model_improvement}.
We then train the off-the-shelf Llama 2 model on these augmented datasets and present both the ROUGE-L scores of the model on the selected domains (refer to ``\qualeval{} Aug.'' columns) and the overall improvement of the ROUGE-L score of the model on the evaluation set (refer to ``$\Delta$ -- Overall'' column).
For the baseline, we use the same training set but randomly augment the training set with $250$ samples (refer to ``No Aug.'' columns).

\subsection{Distance metric for skill usage calibration}
\label{app:distance}

We measure the distance between the affinity scores by measuring the fraction of samples where the difference between the affinity scores of the generation and the ground truth is greater than 1.

\subsection{Example natural language insight}
\input{figures-latex/insights}

\subsection{Dashboards}

\begin{figure*}[t!]
    \centering
    \includegraphics[width=0.9\textwidth]{figures/mbpp/davinci-2/dashboard.pdf}
    \caption{MBPP - \textsc{davinci-2}}
\end{figure*}
\begin{figure*}[t!]
    \centering
    \includegraphics[width=0.9\textwidth]{figures/mbpp/davinci-3/dashboard.pdf}
    \caption{MBPP - \textsc{davinci-3}}
\end{figure*}

\begin{figure*}[t!]
    \centering
    \includegraphics[width=0.9\textwidth]{figures/dialogsum/davinci-2/dashboard.pdf}
    \caption{DialogSum - \textsc{davinci-2}}
\end{figure*}
\begin{figure*}[t!]
    \centering
    \includegraphics[width=0.9\textwidth]{figures/dialogsum/davinci-3/dashboard.pdf}
    \caption{DialogSum - \textsc{davinci-3}}
\end{figure*}

\begin{figure*}[t!]
    \centering
    \includegraphics[width=0.9\textwidth]{figures/mmlu_bio/curie/dashboard.pdf}
    \caption{MMLU (Clinical Knowledge) - \textsc{curie}}
\end{figure*}

\begin{figure*}[t!]
    \centering
    \includegraphics[width=0.9\textwidth]{figures/mmlu_bio/davinci-2/dashboard.pdf}
    \caption{MMLU (Clinical Knowledge) - \textsc{davinci-2}}
\end{figure*}

\begin{figure*}[t!]
    \centering
    \includegraphics[width=0.9\textwidth]{figures/mmlu_bio/davinci-3/dashboard.pdf}
    \caption{MMLU (Clinical Knowledge) - \textsc{davinci-3}}
\end{figure*}

%% file: figures-latex/discovery_prompt.tex
\begin{figure}[ht]
\centering
\begin{adjustbox}{max width=0.9\columnwidth}
\begin{tcolorbox}[colback=intsructionscolor!7!white,colframe=intsructionscolor!90!black,title=MBPP]
\textbf{Domain:} Given the following examples, What are relevant domains for the following programs? Focus on the example programs BUT be general. Structure the response as a numbered list.\\

\textbf{Sub-task:} Given the example programs, What are specific ATOMIC sub-tasks a machine learning model need to be competent at for the underlying task? Focus on the example programs BUT be general. [IMPORTANT] Do NOT list the overall task as a subtask and be GENERAL. Structure the response as: Subtask:. Generate a numbered list.
\end{tcolorbox}
\end{adjustbox}
\begin{adjustbox}{max width=0.9\columnwidth}
    \begin{tcolorbox}[colback=examplescolor!7!white,colframe=examplescolor!75!black,title=DialogSum]
\textbf{Domain:} Given the following conversations, What are relevant domains for the data? Focus on the example data BUT be general. Structure the response as a numbered list. \\

\textbf{Sub-task:} Given the example conversations, What are specific sub-tasks a machine learning model need to be competent at for the underlying task? Focus on the example data BUT be general. [IMPORTANT] Do NOT list the overall task as a subtask and be GENERAL. Structure the response as: Subtask:. Generate a numbered list.

\end{tcolorbox}
\end{adjustbox}

\begin{adjustbox}{max width=0.9\columnwidth}
\begin{tcolorbox}[colback=inputcolor!7!white,colframe=inputcolor!75!white,title=MMLU (Clinical Knowledge)]
\textbf{Domain:} Given the following examples, What are relevant domains for the data? Focus on the example data BUT be general. Structure the response as a numbered list.\\

\textbf{Sub-task:} Given the example questions and answers on clinical biology, What are the sub-tasks a machine learning model needs to be competent at to be a good medical assistant. Focus on the example data BUT please be general. [IMPORTANT] Do NOT list the overall task as a subtask and be GENERAL while being GROUNDED in the example data. Structure the response as: Subtask: $<$subtask$>$. Generate a numbered list.
\end{tcolorbox}
\end{adjustbox}
\caption{Prompt used for discovering attributes across different tasks.}
\label{fig:fewshot_ex2}
\end{figure}

%% file: figures-latex/scoring_prompt.tex
\begin{figure}[ht]
\centering
\begin{adjustbox}{max width=0.9\columnwidth}
\begin{tcolorbox}[colback=domaincolor!7!white,colframe=domaincolor!90!black,title=Domain]

Given the input to a language model, Rate to what degree the input belong to each of the following domains. Rate on a scale of 1-5, with 5 being compeletely belongs and 1 being not belonging at all. [Important] For each domain, format the output as, [Domain 1: $<$domain$>$, Score: $<$score$>$, Evidence: $<$ Evidence for score$>$] [Domain 2: $<$domain$>$, Score: $<$score$>$, Evidence: $<$Evidence for score$>$] [Domain N: $<$domain$>$, Score: $<$score$>$, Evidence: $<$Evidence for score$>$]. [Important] Make sure to include concrete evidence based on the input to JUSTIFY the score. Remember you are an ACCURATE, FAITHFUL, CRITICAL and FAIR judge. \\
\end{tcolorbox}
\end{adjustbox}

\begin{adjustbox}{max width=0.9\columnwidth}
\begin{tcolorbox}[colback=subtaskcolor!7!white,colframe=subtaskcolor!90!black,title=Subtask]
Given the input to a language model, Rate to what degree each of the following subtasks are needed to successfully understand and complete the task. Rate on a scale of 1-5, with 5 being very used and 1 being not used at all. [Important] For each subtask, format the output as [Subtask 1: $<$subtask$>$, Score: $<$score$>$, Evidence: $<$Evidence for score$>$] [Subtask 2: $<$subtask$>$, Score: $<$score$>$, Evidence: $<$Evidence for score$>$] [Subtask N: $<$subtask$>$, Score: $<$score$>$; Evidence: $<$Evidence for score$>$]. [IMPORTANT] Do NOT add $\setminus$n between subtask, score and explanation. [Important] Make sure to include concrete evidence based on the input to JUSTIFY the score. Remember you are an ACCURATE, FAITHFUL, CRITICAL and FAIR judge.
\end{tcolorbox}
\end{adjustbox}
\caption{Prompt for scoring attributes.}
\end{figure}

%% file: figures-latex/insight_prompt.tex
\begin{figure}[ht]
\centering
\begin{adjustbox}{max width=0.9\columnwidth}
\begin{tcolorbox}[colback=maybe!7!white,colframe=maybe!90!black,title=Insight Generation]
\textbf{System:} Given a holistic picture of the performance of a machine learning model, you are asked to summarize the model's overall performance.\\

\textbf{Prompt:} Given the above information, please write a brief summary highlighting important information. Please be precise and concise but please be comprehensive.\\

A machine learning model is tasked with the following task: $\{$ task instruction $\}$\\

These are the {subtasks/domains} for the task: {list of subtasks/domains}\\

In the evaluation data, these are the importance scores of the 
Subtask/Domains: \{json.dumps(prior probabilities of subtasks and domains)\} \\

The following scores show how well the model performs on the subtasks/domains:\{json.dumps(proficiency scores of subtasks and domains)\}\\

The following distance demonstrates how much the domains/subtasks are actually used for generating the output when they are requried to generate the input. Therefore, a low distance implies that the model is utilizing the category when it needs to: \{json.dumps(correlation scores of category)\}. [Important] Lower distance implies the {category} is leveraged when it needs to be used.
\end{tcolorbox}
\end{adjustbox}
\caption{Prompt for generating insights.}
\end{figure}

%% file: figures-latex/insights.tex
\begin{figure}
\begin{adjustbox}{max width=0.9\columnwidth}
\begin{tcolorbox}[colback=inputcolor!7!white,colframe=inputcolor!75!white,title=\qualeval{} Insights]
The machine learning model performs well on various subtasks, with the highest scores in ``Understand test cases'' and ``Validate against test cases''. It also excels in ``Generate Python syntax'' and ``Manage variable assignments and data manipulation''. However, it could improve in ``Implement algorithmic operations'' and ``Handling loops and conditionals''. The model effectively utilizes the subtasks when generating the output, particularly in ``Generate Python syntax'' and ``Implement mathematical operations''. In terms of domains, it performs strongly in ``Sequence Analysis'' and ``String Manipulation'', while improvements can be made in ``Tuple Manipulation'' and ``Number Manipulation''. Overall, the model demonstrates proficiency in understanding the requirements and generating accurate Python code, with potential for further enhancements in specific areas.
\end{tcolorbox}
\end{adjustbox}
\caption{Natural language insights generated by \qualeval{} for the \textit{davinci-2} model on MBPP.}
\label{fig:insights}
\end{figure}